\pdfoutput=1

\documentclass[11pt]{article}

\usepackage{ACL2023}
\usepackage{amsfonts}
\usepackage{times}
\usepackage{latexsym}
\usepackage{makecell}
\usepackage{amsmath}
\usepackage[T1]{fontenc}

\usepackage[utf8]{inputenc}

\usepackage{microtype}

\usepackage{inconsolata}
\usepackage{booktabs}
\usepackage{graphicx}

\usepackage{todonotes}
\usepackage{ulem}

\title{Enhancing Grammatical Error Correction Systems with Explanations}
\author{Yuejiao Fei$^{\spadesuit \clubsuit}$\footnotemark[1]\hspace{0.5mm}, Leyang Cui$^{\heartsuit}$\footnotemark[2]\hspace{0.5mm}, Sen Yang$^{\diamondsuit}$, Wai Lam$^{\diamondsuit}$, Zhenzhong Lan$^{\clubsuit}$\hspace{0.5mm}, Shuming Shi$^{\heartsuit}$ \\ 
 $^\spadesuit$ Zhejiang University\\
  $^\heartsuit$ Tencent AI lab \\
   $^\diamondsuit$ The Chinese University of Hong Kong \\
 $^\clubsuit$ School of Engineering, Westlake University\\
  \quad\texttt{\{feiyuejiao,lanzhenzhong\}@westlake.edu.cn} \\
  \quad\texttt{\{leyangcui,shumingshi\}@tencent.com} \\ 
  \quad\texttt{\{syang, wlam\}@se.cuhk.edu.hk}\\
\\ 
}

\begin{document}
\maketitle
\renewcommand{\thefootnote}{\fnsymbol{footnote}}
\footnotetext[1]{Work was done during the internship at Tencent AI lab.}
\footnotetext[2]{Corresponding authors.}
\renewcommand{\thefootnote}{\arabic{footnote}}
\begin{abstract}
Grammatical error correction systems improve written communication by detecting and correcting language mistakes. To help language learners better understand why the GEC system makes a certain correction, the causes of errors (evidence words) and the corresponding error types are two key factors. To enhance GEC systems with explanations, we introduce EXPECT, a large dataset annotated with evidence words and grammatical error types. We propose several baselines and analysis to understand this task. Furthermore, human evaluation verifies our explainable GEC system's explanations can assist second-language learners in determining whether to accept a correction suggestion and in understanding the associated grammar rule.



\end{abstract}

\section{Introduction}

Grammatical Error Correction (GEC) systems aim to detect and correct grammatical errors in a given sentence and thus provide useful information for second-language learners.
There are two lines of work for building GEC systems. Sequence-to-sequence methods \cite{rothe2021simple, flachs2021data, syngec} take an erroneous sentence as input and generate an error-free sentence autoregressively. Sequence labeling methods \cite{omelianchuk2020gector, tarnavskyi2022ensembling} transform the target into a sequence of text-editing actions and use the sequence labeling scheme to predict those actions.  

With advances in large pre-trained models \cite{devlin2018bert, lewis2019bart} and availability of high-quality GEC corpora \cite{ng2014conll, bryant2019bea}, academic GEC systems \cite{omelianchuk2020gector, rothe2021simple} have achieved promising results on benchmarks and serve as strong backbones for modern writing assistance applications (e.g., Google Docs\footnote{\url{https://www.google.com/docs/about/}}, Grammarly\footnote{\url{https://demo.grammarly.com/}}, and Effidit \cite{effidit}\footnote{\url{https://effidit.qq.com/}}).
Although these academic methods provide high-quality writing suggestions, they rarely offer explanations with specific clues for corrections.
Providing a grammar-aware explanation and evidence words to support the correction is important in second-language education scenarios \cite{ellis2006implicit}, where language learners need to ``know why'' than merely ``know how''.
As a commercial system, Grammarly does provide evidence words, but in very limited cases, and the technical details are still a black box for the research community. 

\begin{figure}[!t]
\includegraphics[width=0.48\textwidth]{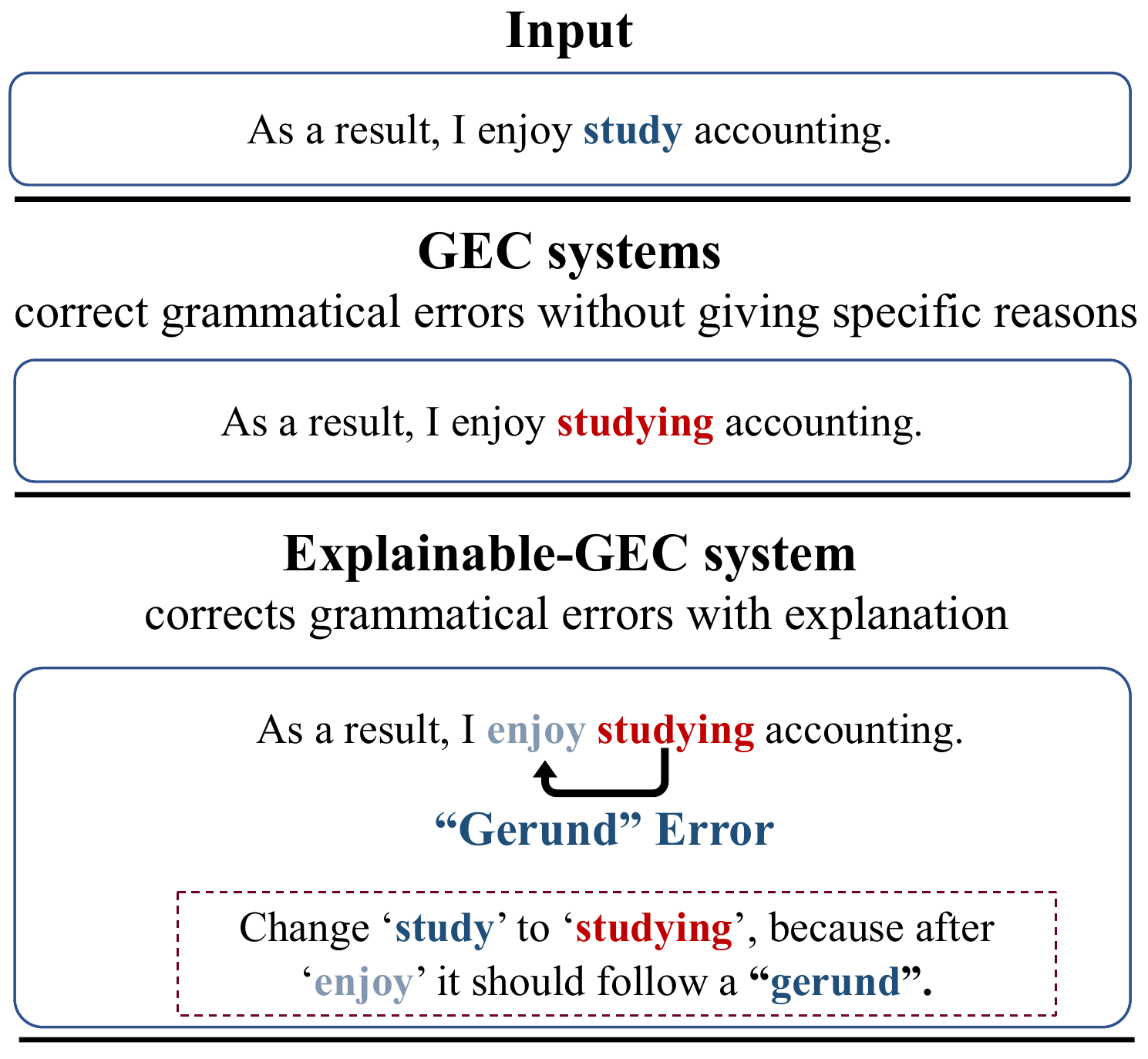}
\caption{Comparison between explainable GEC and conventional GEC systems.}
\label{fig:intro_fig}
\end{figure}
Though some existing work has attempted to enhance the explainability of GEC's corrections \cite{bryant2017automatic, omelianchuk2020gector, kaneko2022interpretability}, they do not provide intra-sentence hints (i.e., evidence words in the sentence).
To fill this gap, we build a dataset named {\bf EXP}lainble grammatical {\bf E}rror {\bf C}orrec{\bf T}ion ({\bf EXPECT}) on the standard GEC benchmark (W\&I+LOCNESS \cite{bryant2019bea}), yielding 21,017 instances with explanations in total.
As shown in Figure~\ref{fig:intro_fig}, given a sentence pair consisting of an erroneous sentence and its corrected counterpart, our explainable annotation includes:
\begin{itemize}
    \item[1)] \textbf{Evidence words} in the erroneous sentence.
        Error tracing could be rather obscure for second-language beginners. 
        For example, given an erroneous sentence, ``\textit{As a result, I enjoy study accounting.}'' where ``\textit{study}'' should be corrected to ``\textit{studying}'', a beginning learner might mistakenly attribute ``\textit{studying}'' to ``\textit{accounting}'' because they both have an ``\textit{ing}'' suffix. However, the correct attribution should be ``\textit{enjoy}''.
        Such incorrect judgment may lead the language learner to draw wrong conclusions (e.g., A verb needs to have an ``\textit{ing}'' suffix if a subsequent verb does so), which significantly disturbs the learning procedure.
        To remedy this, EXPECT provides annotated evidence words, which enable training models to automatically assist second-language learners in finding error clues. 
    \item[2)]
    \textbf{Error types} of the grammatical errors, ranging among the 15 well-defined categories by consulting the pragmatic errors designed by \citet{skehan1998cognitive} and \citet{shichungui}. 
         Language learning consists of both abstract grammar rules and specific language-use examples. A model trained with EXPECT bridges the gap between the two parts: such a model can produce proper error types, automatically facilitating language learners to infer abstract grammar rules from specific errors in an inductive reasoning manner. Further, it allows learners to compare specific errors within the same category and those of different categories, benefiting the learner's inductive and deductive linguistic reasoning abilities.
            
\end{itemize}


To establish baseline performances for explainable GEC on EXPECT, we explore generation-based, labeling-based, and interaction-based methods.
Note that syntactic knowledge also plays a crucial role in the human correction of grammatical errors. For example, the evidence word of subject-verb agreement errors can be more accurately identified with the help of dependency parsing. Motivated by these observations, 
we further inject the syntactic knowledge produced by an external dependency parser into the explainable GEC model.

Experiments show that the interaction-based method with prior syntactic knowledge achieves the best performance (${F}_{0.5}=$70.77).
We conduct detailed analysis to provide insights into developing and evaluating an explainable GEC system.
Human evaluations suggest that the explainable GEC systems trained on EXPECT can help second language learners to understand the corrections better. We will release EXPECT (e.g., baseline code, model, and human annotations) on \url{https://github.com/lorafei/Explainable_GEC}.

\section{Related Work} \label{relatedwork}
\noindent
Some work formulates GEC as a sequence-to-sequence problem. Among them, transformer-based GEC models~\cite{rothe2021simple} have attained state-of-the-art performance on several benchmark datasets~\cite{ng2014conll,bryant2019bea} using large PLMs \cite{t5} and synthetic data \cite{synthetic-data}.
To avoid the low-efficiency problem of seq2seq decoding, some work~\cite{awasthi2019parallel, omelianchuk2020gector, gector-large} formats GEC as a sequence labeling problem and achieves competitive performance.
Both lines of work focus on improving the correction performance and decoding speed but cannot provide users with further suggestions.

Several methods have been proposed to provide explanations for GEC systems.
ERRANT \cite{bryant2017automatic} designs a rule-based framework as an external function to classify the error type information given a correction. GECToR \cite{omelianchuk2020gector} pre-defines g-transformations tag (e.g., transform singular nouns to plurals) and uses a sequence labeling model to predict the tag as explanations directly.
Example-based GEC \cite{kaneko2022interpretability} adopts the k-Nearest-Neighbor method \cite{khandelwal2019generalization} for GEC, which can retrieve examples to improve interpretability.
Despite their success, their explanations are restricted by pre-defined grammar rules or unsupervised retrieval.
They may not generalize well to real-life scenarios due to the limited coverage of the widely varying errors made by writers.  
In contrast, our annotated instances are randomly sampled from real-life human-written corpora without restriction, thus providing a much larger coverage.

\citet{nagata2019toward, nagata2020creating, hanawa2021exploring}, and \citet{nagata2021shared} propose a feedback comment generation task and release two corresponding datasets, which, to our knowledge, are the only two publicly available and large-scale datasets focusing on GEC explanations.
The task aims to generate a fluent comment describing the erroneous sentence's grammatical error. 
While this task integrates GEC and Explainable-GEC into one text generation task, 
we only focus on Explainable-GEC and formulate it as a labeling task, which is easier and can avoid the high computational cost of seq2seq decoding.
Furthermore, the evaluation of feedback comment generation mainly relies on human annotators to check if the error types are correctly identified and if the grammatical error correction is proper in the generated text, which is time-consuming and susceptible to the variations resulting from subjective human judgment.
In contrast, our token classification task can be easily and fairly evaluated by automatic metrics (e.g., F-score), favoring future research in this direction.




\section{Dataset}

\begin{figure*}[!t]
\centering
\includegraphics[width=1.8\columnwidth]{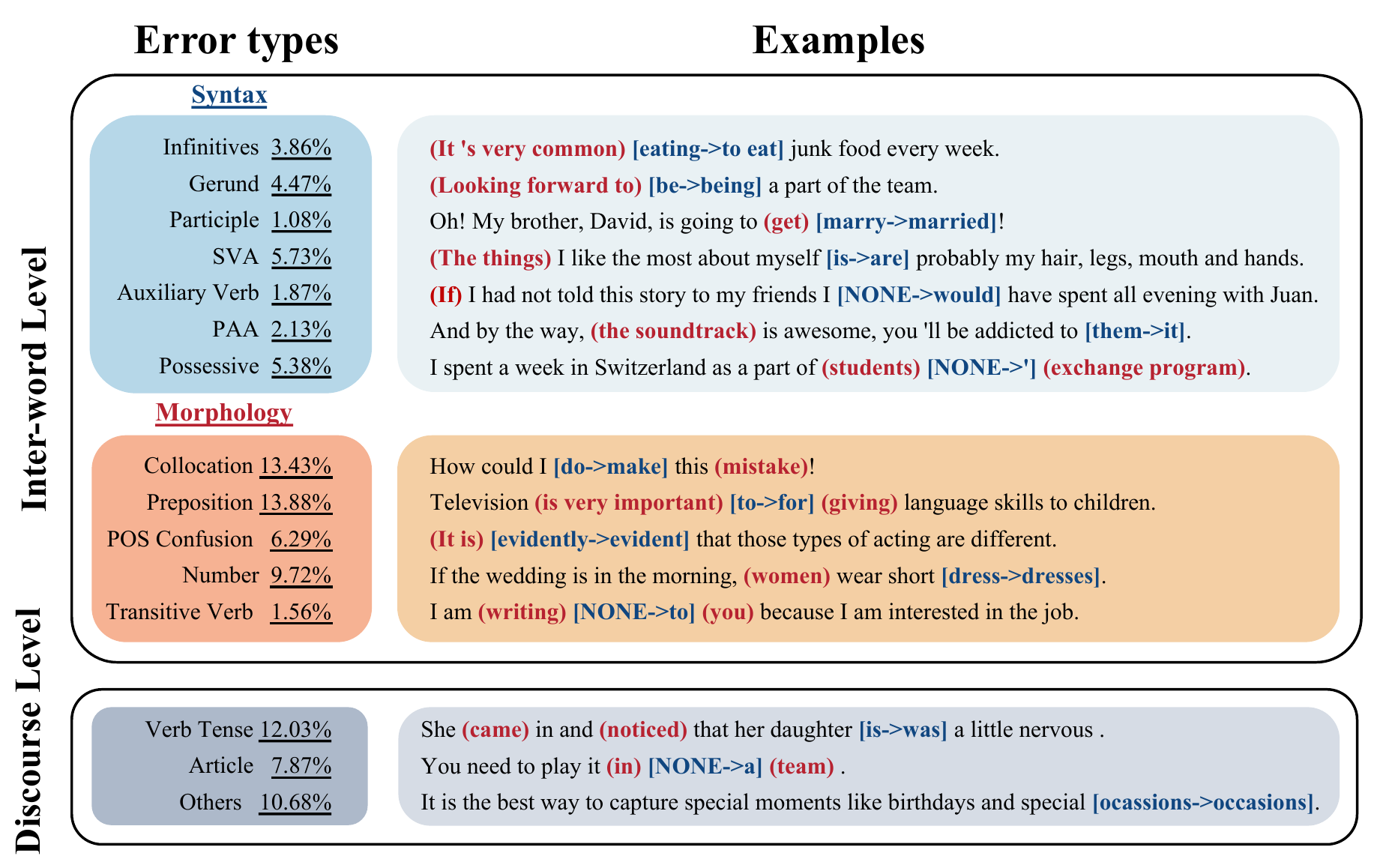}
\caption{Examples of each error type and corresponding evidence words in EXPECT. Blue text indicates the correction, while red text indicates the evidence words. SVA means subject-verb agreement, PAA means pronoun-antecedent agreement, POS confusion means part-of-speech confusion.}
\label{fig:overview}
\end{figure*}

To facilitate more explainable and instructive grammatical error correction,
we propose the EXPECT, an English grammatical error explanation dataset annotated with 15 grammatical error types and corresponding evidence words.

\subsection{Data Source}
We annotate EXPECT based on W\&I+LOCNESS \cite{bryant2019bea}, which comprises 3,700 essays written by international language learners and native-speaking undergraduates and corrected by English teachers. 
We first select all sentences with errors from essays. For a sentence with $n$ errors, we repeat the sentence $n$ times and only keep a single unique error in each sentence. Then, we randomly sample and annotate 15,187 instances as our training set. We do the same thing for the entire W\&I+LOCNESS dev set, and split it up into test and development sets evenly. 

In order to better align with real-world application scenarios, we have additionally annotated 1,001 samples based on the output of the conventional GEC models. We randomly sampled the output of T5-large \cite{rothe2021simple} and GECToR-Roberta \cite{omelianchuk2020gector} on the W\&I+LOCNESS test set. We also report whether the corrections of the GEC model were right. 

\subsection{Error Type Definition}

Following the cognitive model of second language acquisition \cite{skehan1998cognitive, shichungui}, we design error types among three cognitive levels as follows:
 
\textbf{Single-word level error} is in the first and lowest cognitive level. These mistakes usually include misuse of \textit{spelling}, \textit{contraction}, and \textit{orthography}, which are often due to misremembering. Since there is no clear evidence for those errors, we classify them into type \textit{others}. 

\textbf{Inter-word level error} is in the second cognitive level, which usually stems from a wrong understanding of the target language. Most error types with clear evidence lie at this level because it represents the interaction between words. This level can be further split into two linguistic categories, syntax class and morphology class: 
(1) In the view of syntax, we have seven error types, including \textit{infinitives}, \textit{gerund}, \textit{participles}, \textit{subject-verb agreement}, \textit{auxiliary verb}, \textit{pronoun} and \textit{noun possessive}. 
(2) In the view of morphology, we have five error types, including \textit{collocation}, \textit{preposition}, \textit{word class confusion}, \textit{numbers}, and \textit{transitive verbs}. 

\textbf{Discourse level error} is at the highest cognitive level, which needs a full understanding of the context. 
These errors include \textit{punctuation}, \textit{determiner}, \textit{verb tense}, \textit{word order} and \textit{sentence structure}. 
Since \textit{punctuation}, \textit{word order}, and \textit{sentence structure} errors have no clear evidence words, we also classify them into type \textit{others}.

The complete list of error types and corresponding evidence words are listed in Figure \ref{fig:overview}.  The definition of each category is shown in Appendix \ref{sec:appendixa}.

\subsection{Annotation Procedure}
\label{sec:human_annotatation}
Our annotators are L2-speakers who hold degrees in English and linguistics, demonstrating their proficiency and expertise in English. The data are grouped into batches of 100 samples, each containing an erroneous sentence and its correction. The annotators are first trained on labeled batches until their ${F}_{1}$ scores are comparable to those of the main author. After that, annotators are  asked to classify the type of the correction and highlight \textit{evidence words} that support this correction on the unlabeled batches. The evidence should be \textbf{informative} enough to support the underlying grammar of the correction meanwhile \textbf{complete} enough to include all possible evidence words. For each complete batch, we have an experienced inspector to re-check 10\% of the batch to ensure the annotation quality. According to inspector results, if ${F}_{1}$ scores for the annotation are lower than 90\%, the batch is rejected and assigned to another annotator.



\subsection{Data Statistics}
The detailed statistics of EXPECT have listed in Table \ref{table:stats}.
Take the train set for example, the average number of words per sentence is $28.68$, and $74.15\%$ of the entire dataset has explainable evidence.
Among all sentences with evidence words, the average number of words per evidence is $2.59$.
The percentage of all error types is listed in Figure \ref{fig:overview}. Detailed description for all error categories is listed in Appendix~\ref{sec:appendixa}. 
Top-3 most frequent error types are \textit{preposition} (13.88\%), \textit{collocation} (13.43\%) and \textit{verb tense} (12.03\%). 

\begin{table}[!t]
\centering
\small
\resizebox{0.95\columnwidth}!{\begin{tabular}{l|cccc}
\hline
\textbf{Data Statistics} & \textbf{Train} & \textbf{Dev} & \textbf{Test} & \textbf{Outputs}  \\ \hline
Number of sentences          & 15,187   & 2,413    & 2,416  & 1001  \\
Number of words              & 435,503  & 70,111   & 70,619 & 27,262  \\
Avg. w.p.s                   & 28.68    & 29.06    & 29.23  & 27.23   \\
With evidence rate           & 74.15    & 59.10    & 59.77  & 72.73 \\
Total evidence words         & 29,187   & 4,280    & 4,340  & 1736  \\
Avg. evidence w.p.s          & 2.59     & 3.00     & 3.01   & 2.38  \\ \hline
\end{tabular}}
\caption{Data Statistics of EXPECT. Here w.p.s means word per sentence.}
\label{table:stats}
\end{table}

\subsection{Evaluation Metrics}
We consider our task as a token classification task. Thus we employ token-level (precision, recall, $F_\text{1}$, and $F_\text{0.5}$) and sentence-level (exact match, label accuracy) evaluation metrics. Specifically, the \textbf{exact match} requires identical error types and evidence words between label and prediction, and the \textbf{label accuracy} measures the classification performance of error types. To explore which automatic metric is more in line with human evaluation, we compute Pearson correlation \cite{freedman2007statistics} between automatic metrics and human judgment. As shown in Table \ref{tab:correlation}, $F_\text{0.5}$ achieves the highest score in correlation. And precision is more correlated with human judgment than recall. The reason may be that finding the precise evidence words is more instructive than extracting all evidence words for explainable GEC.

\begin{table}[!t]
\centering
\resizebox{0.85\columnwidth}!{\begin{tabular}{c|c|c|c|c}
\hline
\textbf{Precision} & \textbf{Recall} & $\mathbf{F}_{1}$ & $\mathbf{F}_{0.5}$ & \textbf{Exact Match} \\
\hline
0.469 & 0.410 & 0.463 & \textbf{0.471} & 0.342 \\
\hline
\end{tabular}}
\caption{Pearson correlation between human judgment and different automatic evaluation metrics.}
\label{tab:correlation}
\end{table}

\begin{figure*}[!t]
\centering
\includegraphics[width=2\columnwidth]{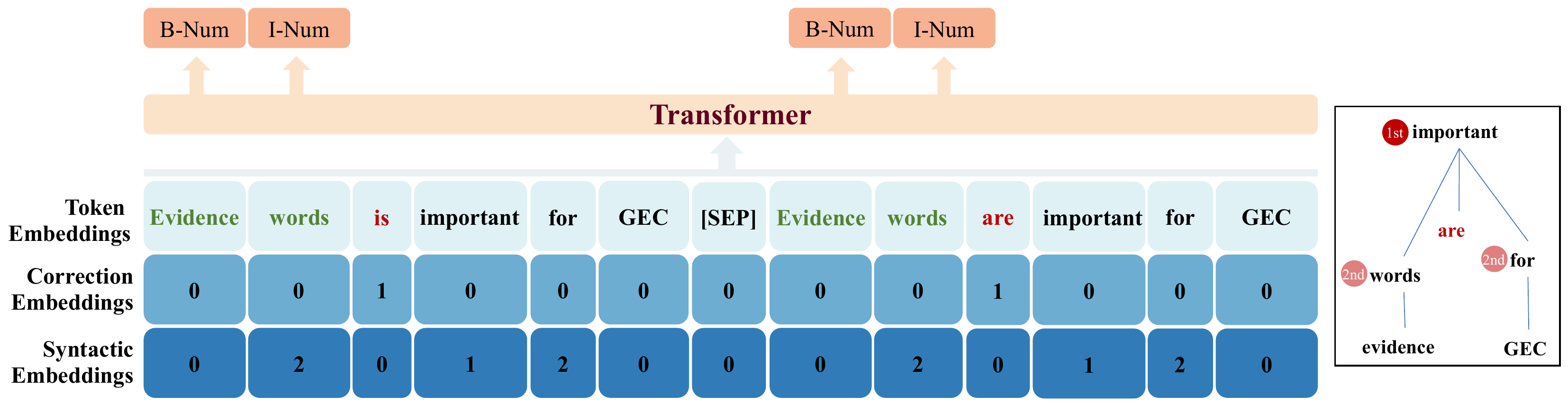}
\caption{An illustration of labeling-based methods with syntax for solving explainable GEC. On the right is the dependency parsing tree of the corrected sentence, where the correction word \textit{are} is marked in red, and 1st and 2nd-order nodes are marked with red circles.}
\label{fig:model}
\end{figure*}
\section{Methods}

In this section, we define the task of explainable-GEC in Section~\ref{sec:method:task-formulation} and then introduce the labeling-based baseline method in Section~\ref{sec:method:span}, and the interaction method in Section~\ref{sec:method:interactive}.

\subsection{Task Formulation}
\label{sec:method:task-formulation}

The task input is a pair of sentences, including an erroneous sentence $X = \{ x_1, x_2, ..., x_n \}$ and the corresponding corrected sentence $Y = \{ y_1, y_2, ..., y_m \}$.
The two sentences usually share a large ratio of overlap.
The difference between the two sentences is defined as a span edit $\{(s_x, s_y)\}$. 
The task of explainable GEC is to find the grammar evidence span $E_x$ within $X$ and predict corresponding error type classes $c$.
Take Figure~\ref{fig:model} as an example, $s_x = \text{ ``are'' and } s_y = \text{ ``is''}$, the evidence span $E_x=$``Evidence words''. 

\subsection{Labeling-based Method}
\label{sec:method:span}
We adopt the labeling-based method for explainable GEC.

\paragraph{Input.}
We concatenate the erroneous sentence $X$ and the corresponding error-free sentence $Y$, formed as $\texttt{[CLS]} X \texttt{[SEP]} Y \texttt{[SEP]}$.

\paragraph{Correction Embedding.} 
To enhance the positional information of the correction, we adopt a correction embedding $e_c$ to encode the position of the correction words in the sentence $X$ and $Y$. We further add $\mathbf{e}_c$ to embeddings in BERT-based structure as follow: 
\begin{equation}
    \mathbf{e} = \mathbf{e}_{t} + \mathbf{e}_{p} + \mathbf{e}_{c}
\end{equation}
where $\mathbf{e}_{t}$ is the token embeddings, and $\mathbf{e}_{p}$ is the position embeddings.

\paragraph{Syntactic Embedding.}
There is a strong relation between evidence words and syntax as shown in Section \ref{sec:syntax}. Hence we inject prior syntactic information into the model. 
Firstly, given the corrected sentence $Y$ and its span edit $(s_x,s_y)$, we parse sentence $Y$ with an off-the-shell dependency parser from the AllenNLP library \cite{gardner2018allennlp}. 
For each word in $s_y$, we extract its first-order dependent and second-order dependent words in the dependency parse tree. For example, as shown in Figure~\ref{fig:model}, the correction word $s_y =$ ''are'', the first-order dependent word is ''important'', and the second-order dependent words are ''words'', and ''for'', and they are marked separately. By combining all correction edits' first-order words and second-order words, we construct the syntactic vector $d_Y \in \mathbb{R}^m$ for sentence $Y$.
Dependency parsing is originally designed for grammatical sentences. To acquire the syntax vector of the erroneous sentence $X$, we use the word alignment to map the syntax-order information from the corrected sentence to the erroneous sentence, yielding $d_X \in \mathbb{R}^n$. 
We then convert $[d_X, d_Y]$ to syntactic embedding $\mathbf{e}_s$, and add to the original word embedding:
\begin{equation}
    \mathbf{e} = \mathbf{e}_{t} + \mathbf{e}_{p} + \mathbf{e}_{c} + \mathbf{e}_{s}
\end{equation}

\paragraph{Encoder.} 
We adopt a pre-trained language model (e.g. BERT) as an encoder to encode the input $\mathbf{e}$, yielding a sequence of hidden representation $\mathbf{H}$.

\paragraph{Label Classifier.} The hidden representation $\mathbf{H}$ is fed into a classifier to predict the label of each word. The classifier is composed of a linear classification layer with a softmax activation function. 

\begin{equation}
\hat{l}_i =  \mathrm{softmax}(\mathbf{W} \mathbf{h}_i + \mathbf{b}),
\end{equation}where $\hat{l}_i$ is the predicted label for $i$-th word, $\mathbf{W}$ and $\mathbf{b}$ are the parameters for the softmax layer.

\paragraph{Training.} The model is optimized by the log-likelihood loss. For each sentence, the training object is to minimize the cross-entropy between $l_i$ and $\hat{l_i}$ for a labeled gold-standard sentence. 
\begin{equation}
\mathcal{L} = - \sum_{i}^{m+n+1} \log \hat{l}_i.
\end{equation}

\begin{figure}[!t]
\centering
\includegraphics[width=0.9\columnwidth]{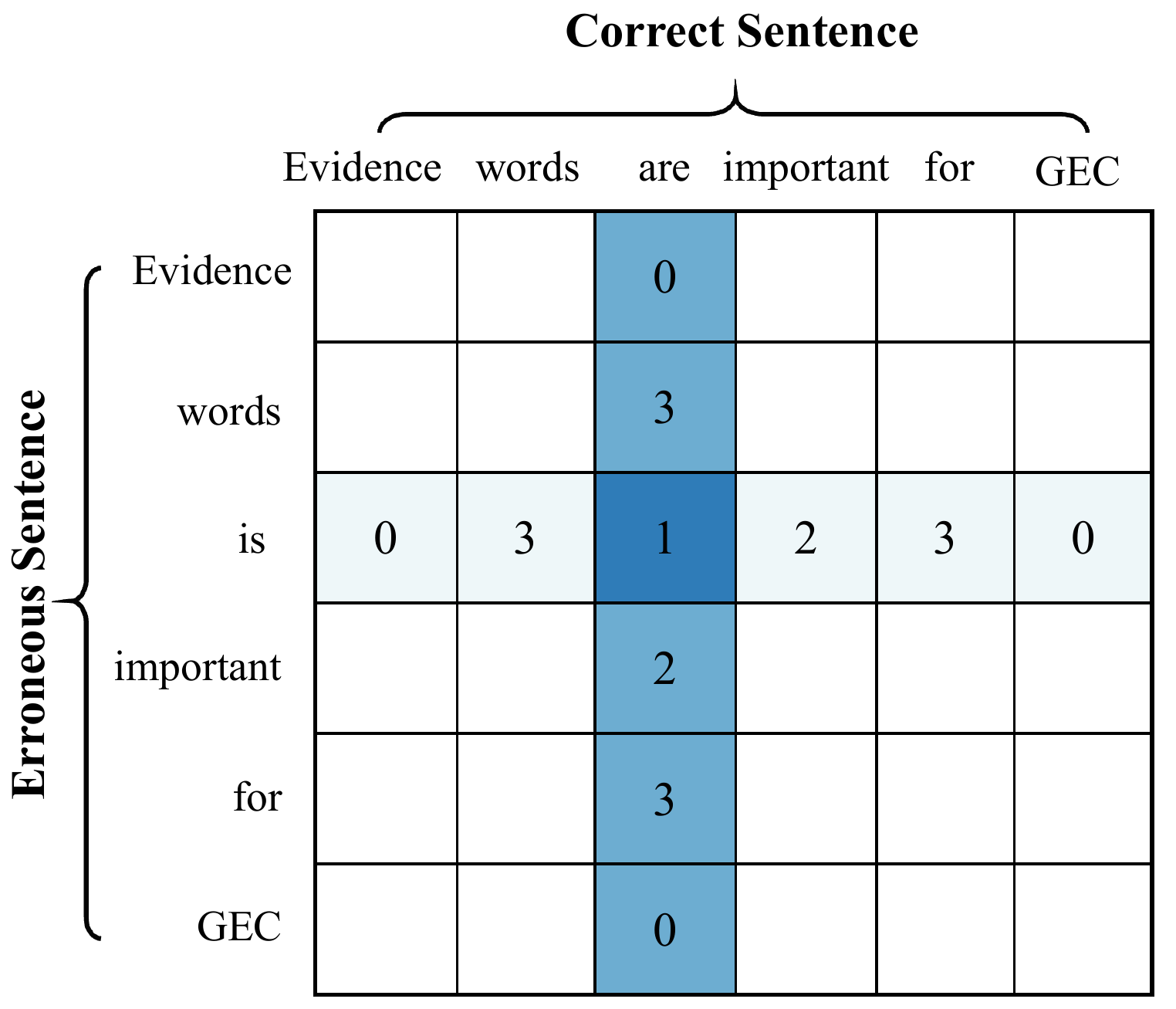}
\caption{Syntactic Interactive Matrix.}
\label{fig:parsing_matrix}
\end{figure}

\subsection{Interaction-based Method} 
%


\label{sec:method:interactive}
Although labeling-based methods model the paired sentences in a joint encoder, it still predicts two separate outputs independently.
The dependencies between the erroneous sentence and the corrected sentence are not explicitly modeled.
Intuitively, the alignment between the erroneous sentence and the corrected sentence can be highly informative.
We propose an interactive matrix to jointly model the alignment and the evidence span.  
In particular, we adopt a bi-affine classifier to model the multiplicative interactions between the erroneous sentence and the corrected sentence.
Assume that the hidden representation of the erroneous sentence and the corrected sentence are $\mathbf{H}^e$ and $\mathbf{H}^c$, respectively.

We first use two separate feed-forward networks to map the hidden representation into an erroneous query representation and a corrected key representation:
\begin{equation}
\begin{split}
       \mathbf{H}^q &= \mathbf{W}^q \mathbf{H}^e + \mathbf{b}^e \\
    \mathbf{H}^k &= \mathbf{W}^k \mathbf{H}^c + \mathbf{b}^c
\end{split}
\end{equation}

Then a bi-affine attention \cite{biaffine} is adopted to model the interaction between $\mathbf{H}^q$ and $\mathbf{H}^k$:
\begin{equation}
    \mathbf{\hat{M}} = softmax(\mathbf{H}^q \mathbf{U} \mathbf{H}^k + \mathbf{b}^U),
\label{eq:matrix}
\end{equation}
where $\mathbf{U} \in \mathbb{R}^{|H|\times|H|\times|L|}$, $|H|$ and $|L|$ indicates the hidden size and the size of the label set.

\paragraph{Training.}
Similar to the labeling-based method, the training objective is to minimize the cross-entropy between $\mathbf{M}$ and $\mathbf{\hat{M}}$ given a labeled gold-standard sentence:
\begin{equation}
\mathcal{L} = - \sum_{i}^{m} \sum_{j}^n \log \hat{M}_{ij}.
\end{equation}





\paragraph{Syntactic Interactive Matrix.} 
Similar to \textit{Syntactic Embedding}, we use a syntactic interactive matrix to better merge the syntactic knowledge into the model. We construct the syntactic interactive matrix $\mathbf{D}^{syn}$ in the same way as the syntactic embedding above, except for using a interactive matrix rather than a flat embedding. Figure~\ref{fig:parsing_matrix} shows an example of a syntactic matrix, where the row of the correction index in the erroneous sentence is placed with a syntactic vector of the corrected sentence, whereas the column of the correction index in a corrected sentence is placed with erroneous sentence's syntactic vector. Then a two-layer MLP is used to map $\mathbf{D}^{syn}$ to $\mathbf{H}^{syn}$:
\begin{equation}
\small
    \mathbf{H}^{syn} =    \mathbf{W}^{\mathrm{syn}}_2  \textsc{ReLU}(\mathbf{W}^{\mathrm{syn}}_1 \mathbf{D}^{syn} + \mathbf{b}^{\mathrm{syn}}_1) + \mathbf{b}^{\mathrm{syn}}_2
\end{equation}
$\mathbf{H}^{syn}$ is then used as an auxiliary term to calculate the interaction matrix $\mathbf{M}$. Eq~\ref{eq:matrix} is reformulated as: 
\begin{equation}
\small
    \mathbf{\hat{M}} = softmax(\mathbf{H}^q \mathbf{U} \mathbf{H}^k + \mathbf{H}^{syn} + \mathbf{b}^U).
\label{eq:syn_matrix}
\end{equation}

\section{Experiments}
\begin{table*}[]
\centering
\resizebox{1.8\columnwidth}{!}{
\begin{tabular}{c|cccc|cc|cccc|cc}
\hline
 & \multicolumn{6}{c|}{\textbf{Dev}} & \multicolumn{6}{c}{\textbf{Test}}   \\ 
\cline{2-13} 
\multicolumn{1}{c|}{\textbf{Methods}} & \multicolumn{1}{c}{\textbf{P}} & \multicolumn{1}{c}{\textbf{R}} & \multicolumn{1}{c}{$\textrm{F}_{1}$} & \multicolumn{1}{c|}{$\textrm{F}_{0.5}$} & \multicolumn{1}{c}{\textbf{EM}} & \multicolumn{1}{c|}{\textbf{Acc}} & \multicolumn{1}{c}{\textbf{P}} & \multicolumn{1}{c}{\textbf{R}} & \multicolumn{1}{c}{$\textrm{F}_{1}$} & \multicolumn{1}{c|}{$\textrm{F}_{0.5}$} & \multicolumn{1}{c}{\textbf{EM}} & \multicolumn{1}{c}{\textbf{Acc}} \\ \hline
\multicolumn{1}{r|}{\textbf{Human}} & - & - & - & -&-&- & 77.50	& 75.98	& 76.73	& 77.19	 & 69.00	& 87.00 \\
\hline  
\multicolumn{1}{r|}{\textbf{Generation-based}} & & & & & & & & & & \\
\multicolumn{1}{r|}{BART-large} & 65.75 & 62.16 & 63.91 & 65.00 & 49.73 & 75.96 & 65.68 & \textbf{61.98} & 63.78 & 64.90 & 49.20 & 79.12 \\
\hline
\multicolumn{1}{r|}{\textbf{Labeling-based}} & & & & & & & & & & \\
\multicolumn{1}{r|}{Error only} & 50.39 & 33.41 & 40.18 & 45.74 & 39.77 & 56.13 & 50.31 & 35.07 & 41.33 & 46.29 & 39.68 & 56.06 \\
\multicolumn{1}{r|}{Correction only} & 24.77 & 14.07 & 17.94 & 21.50 & 29.34 & 37.34 & 23.14 & 12.53 & 16.26 & 19.79 & 28.97 & 37.67 \\
\multicolumn{1}{r|}{Error+Appendix} & 62.92 & 58.36 & 60.55 & 61.95 & 47.85 & 72.33 & 64.78 & 60.81 & 62.73 & 63.94 & 47.91 & 73.27  \\
\multicolumn{1}{r|}{Correction+Appendix} & 64.85 & 55.74 & 59.95 & 62.80 & 50.00 & 74.36 & 61.86 & 54.45 & 57.92 & 60.22 & 47.66 & 72.98 \\
\multicolumn{1}{r|}{Error+Correction} & 67.82 & 57.51 & 62.24 & 65.47 & 50.60 & 72.42 & 68.91 & 57.94 & 62.95 & 66.39 & 59.19 & 77.31 \\
\multicolumn{1}{r|}{Error+Correction+CE} & 69.76 & 62.20 & 65.77 & 68.11 & 54.09 & 75.65 & 69.44 & 60.93 & \textbf{64.91} & 67.55 & 61.39 & 79.14 \\
\multicolumn{1}{r|}{Error+Correction+CE+Syntax} & 70.06 & \textbf{62.44} & \textbf{66.03} & 68.39 & 55.21 & 76.57 & 68.23 & 61.23 & 64.54 & 66.71 & 61.26 & 78.93 \\ \hline
\multicolumn{1}{r|}{\textbf{Interaction-based}} & & & & & & & & & & \\
\multicolumn{1}{r|}{Error+Correction+CE} & 71.63 & 59.54 & 65.03 & 68.83 & 63.04 & 80.05 & 68.47 & 59.14 & 63.46 & 66.38 & 66.28 & 81.17 \\
\multicolumn{1}{r|}{Error+Correction+CE+Syntax} & \textbf{74.77} & 58.31 & 65.52 & \textbf{70.77} & \textbf{64.58} & \textbf{81.34} & \textbf{73.05} & 56.45 & 63.69 &  \textbf{68.99} & \textbf{67.81} & \textbf{81.79} \\ \hline
\end{tabular}}
\caption{Model performance on EXPECT. EM means Exact Match, CE means correction embeddings.}
\label{tab:main_results}
\end{table*}

\subsection{Baseline Methods}
\textbf{Human} performance is reported. We employ three NLP researchers to label the test set and report the average score as human performance. \\
\textbf{Generation-based} method frames the task as a text generation format. It utilizes a pre-trained generation model to predict the type of error and generate a corrected sentence with highlighted evidence words marked by special tokens. \\
\textbf{Labeling-based} (\textbf{\textit{error only}}) method uses only erroneous sentences as input and predicted explanation directly. \\
\textbf{Labeling-based} (\textbf{\textit{correction only}}) method uses only corrected sentences as input and predicted explanation directly. \\
\textbf{Labeling-based} (\textbf{\textit{with appendix}}) method uses only erroneous sentences or corrected sentences and appends correction words at the end of the sentence. \\
\textbf{Labeling-based} (\textbf{\textit{error and correction}}) method concatenate erroneous and corrected sentences as described in Section \ref{sec:method:span}. \\


\subsection{Main Results}
The model performance under different settings are shown in Table \ref{tab:main_results}. 

We evaluate the model performance across a variety of settings, including generation-based, labeling-based, and interaction-based, as well as syntactic-based and non-syntactic-based.
First, we find that generation-based methods do not outperform labeling-based methods and suffer from poor inference efficiency due to auto-regressive decoding.
In addition, interaction-based methods exhibit higher precision but lower recall compared to labeling-based methods.
This is likely due to the interaction between two sentences helping the model identify more evidence words.
Based on labeling-based methods, adding syntactic information has a marginal 0.28 $F_\text{0.5}$ point increase, while for interaction-based methods, the performance increases by 1.94 $F_\text{0.5}$ point. This suggests that syntactic information can generally provide an indication for identifying evidence words. And the interaction matrix better incorporates syntactic information into the model. Particularly, we found correction embeddings are pretty important for this task. With correction embeddings, the performance increases by 2.64 $F_\text{0.5}$ points on Dev set and 1.16 points on Test set. Finally, interaction-based methods with syntactic knowledge achieve the best performance when measured by precision, $F_\text{0.5}$, exact match, and accuracy. 

\subsection{Impact of Syntactic Knowledge}
\label{sec:syntax}
To further explore the role of syntactic knowledge in boosting the explainable GEC performance,
we first analyze the relation between evidence words and correction words' adjacent nodes in the dependency parsing tree. 
As shown in Table \ref{table:ev_train},  46.71\% of instances have at least one evidence word within correction words' first-order nodes, and 27.02\% of instances' all evidence words stay within second-order nodes. We can infer that syntactic knowledge can in a way narrow the search space of extracting evidence words. 
\begin{table}[!t]
\centering
\resizebox{0.66\columnwidth}!{\begin{tabular}{c|c|c}
\hline
 & \textbf{Count} & \textbf{Ratio} \\
\hline
\thead{Exist evidence word in 1st} & 7,094 & 46.71 \\ \hline
\thead{Exist evidence word in 2st}  & 7,723 & 50.85 \\ \hline
\thead{All evidence words in 1st} & 2,528 & 16.65  \\ \hline
\thead{All evidence words in 2st} & 4,103 & 27.02 \\ \hline

\end{tabular}}
\caption{Statistics of training set evidence words within first-order and second-order nodes.}
\label{table:ev_train}
\end{table}

\begin{figure}[!t]
\centering
\includegraphics[width=0.75\columnwidth]{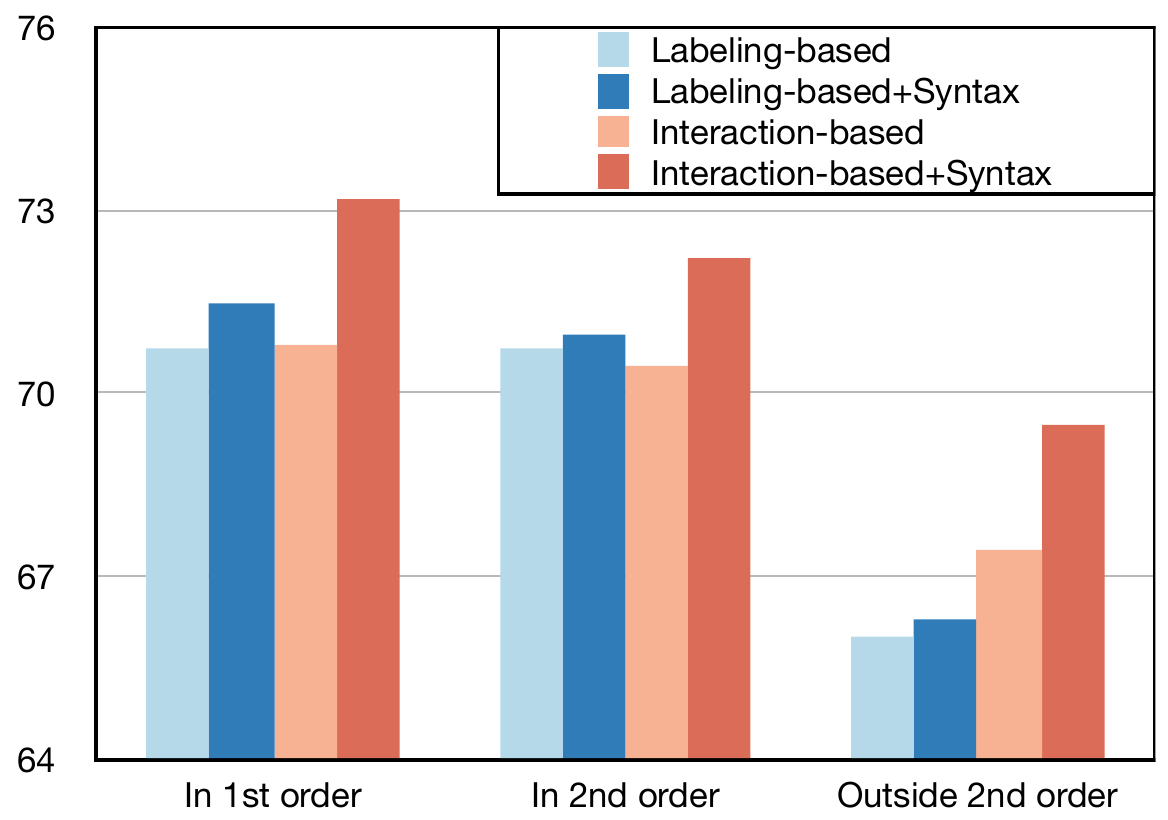}
\caption{$F_\text{0.5}$ score comparison of evidence words in first and second order nodes.}
\label{fig:parsing_order_performance}
\end{figure}

\paragraph{Model Performance across Syntactic Distance.} We compare $F_\text{0.5}$ scores for instances whose evidence words are in and out of the 1st and 2nd dependent orders in Figure \ref{fig:parsing_order_performance}.
The overall performance decreases when evidence words are outside the 2nd dependent order, indicating that the model has trouble in handling complex syntactic structure. But after injecting the syntactic knowledge, the performance increases in all sections, suggesting the effectiveness of syntactic representation.  

\begin{figure}[!t]
\centering
\includegraphics[width=0.7\columnwidth]{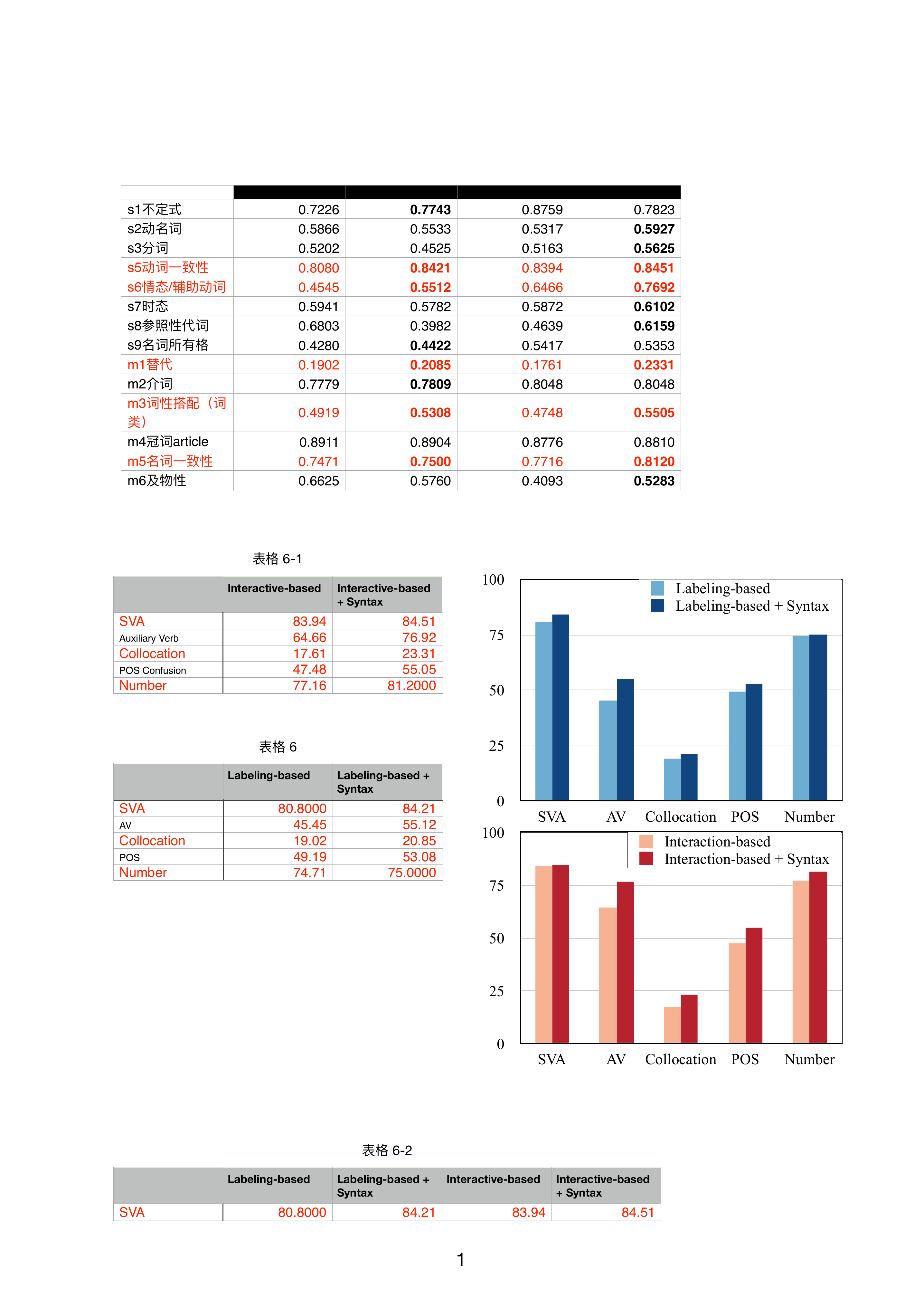}
\caption{$F_\text{0.5}$ score comparison of syntax-related error types between syntactic methods and non-syntactic methods. POS - POS Confusion.}
\label{fig:add_parsing_type}
\end{figure}

\paragraph{Benefit of Syntactic Representation.} We report $F_\text{0.5}$ scores on specific error types before and after injecting syntactic information into the models in Figure \ref{fig:add_parsing_type}. Dependency parsing is a common tool to detect \textit{SVA}\cite{sun2007mining}. The performance on \textit{SVA} indeed increases with the syntax. We also find four other error types which are closely associated with syntactic information, including \textit{auxiliary verb}, \textit{collocation}, \textit{POS confusion} and \textit{number}, whose performance increases significantly for both the labeling-based method and the interaction-based method.

\subsection{Impact of Sentence Length}
Table~\ref{tab:seq_length} illustrates the model performance across different lengths of erroneous sentences. As the sentence length increases, the performance of all methods decreases significantly, which is consistent with human intuition. Longer sentences may contain more complex syntactic and semantic structures, which are challenging for models to capture.

\begin{table}[!t]
\centering
\resizebox{1\columnwidth}!{\begin{tabular}{c|ccccc}
\hline
\thead{\textbf{Sentence} \\\textbf{length}} &  \#\textbf{Samples} & \thead{\textbf{Labeling}\\\textbf{-based}} & \thead{\textbf{Labeling-based} \\ \textbf{+ Syntax}} & \thead{\textbf{Interactive}\\\textbf{-based}} & \thead{\textbf{Interaction-based} \\ \textbf{+ Syntax}} \\
\hline
Less than 10 & 160 & 72.15 & 73.52 & 71.43 & \textbf{77.57} \\
10 to 20 & 751 & 70.44 & 69.96 & 68.22 & \textbf{71.09} \\
20 to 30 & 730 & 67.57 & 67.17 & 70.27 & \textbf{71.68} \\
30 to 40 & 376 & 66.86 & \textbf{69.63} & 67.25 & 69.28 \\
40 to 60 & 239 & 66.86 & 66.88 & 67.01 & \textbf{68.80} \\
More than 60 & 157 & 62.45 & 62.25 & \textbf{70.36} & 64.47 \\
\hline
\end{tabular}}
\caption{Model performance ${F}_{0.5}$ scores across sentence length.}
\label{tab:seq_length}
\end{table}

\subsection{Result on Real-world GEC System}
We employ the gold correction as the input during both the training phase and the inference phase. However, in a practical scenario, this input would be replaced with the output of a GEC system. To evaluate the performance of the explainable system equipped with real-world GEC systems, we use interaction-based methods with syntactic knowledge trained on EXPECT, and directly test using samples that are annotated from the outputs of the GEC model on the W\&I+LOCNESS test set.
The $F_\text{0.5}$ scores obtained are 57.43 for T5-large outputs and 60.10 for GECToR-Roberta outputs, which significantly underperforms 68.39. This may be caused by the training-inference gap as mentioned and the error propagation of the GEC system.

\subsection{Human Evaluation}
\label{sec:human_eval}
To assess the effectiveness of the explainable GEC for helping second-language learners understand corrections, we randomly sample 500 instances with gold GEC correction and 501 outputs decoded by an off-the-shelf GEC system GECTOR \cite{omelianchuk2020gector}, and predict their evidence words and error types using the interaction-based model with syntactic knowledge. We recruit 5 second-language learners as annotators to evaluate whether the predicted explanation is helpful in understanding the GEC corrections. The results show that 84.0 and 82.4 percent of the model prediction for gold GEC correction and GECTOR has explanations, and 87.9 and 84.5 percent of the explanations of EXPECT and gold GEC correction, respectively, are helpful for a language learner to understand the correction and correct the sentence. This show that the explainable GEC system trained on EXPECT can be used as a post-processing module for the current GEC system.

\subsection{Case Study}
\begin{figure}[!t]
\centering
\includegraphics[width=1.0\columnwidth]{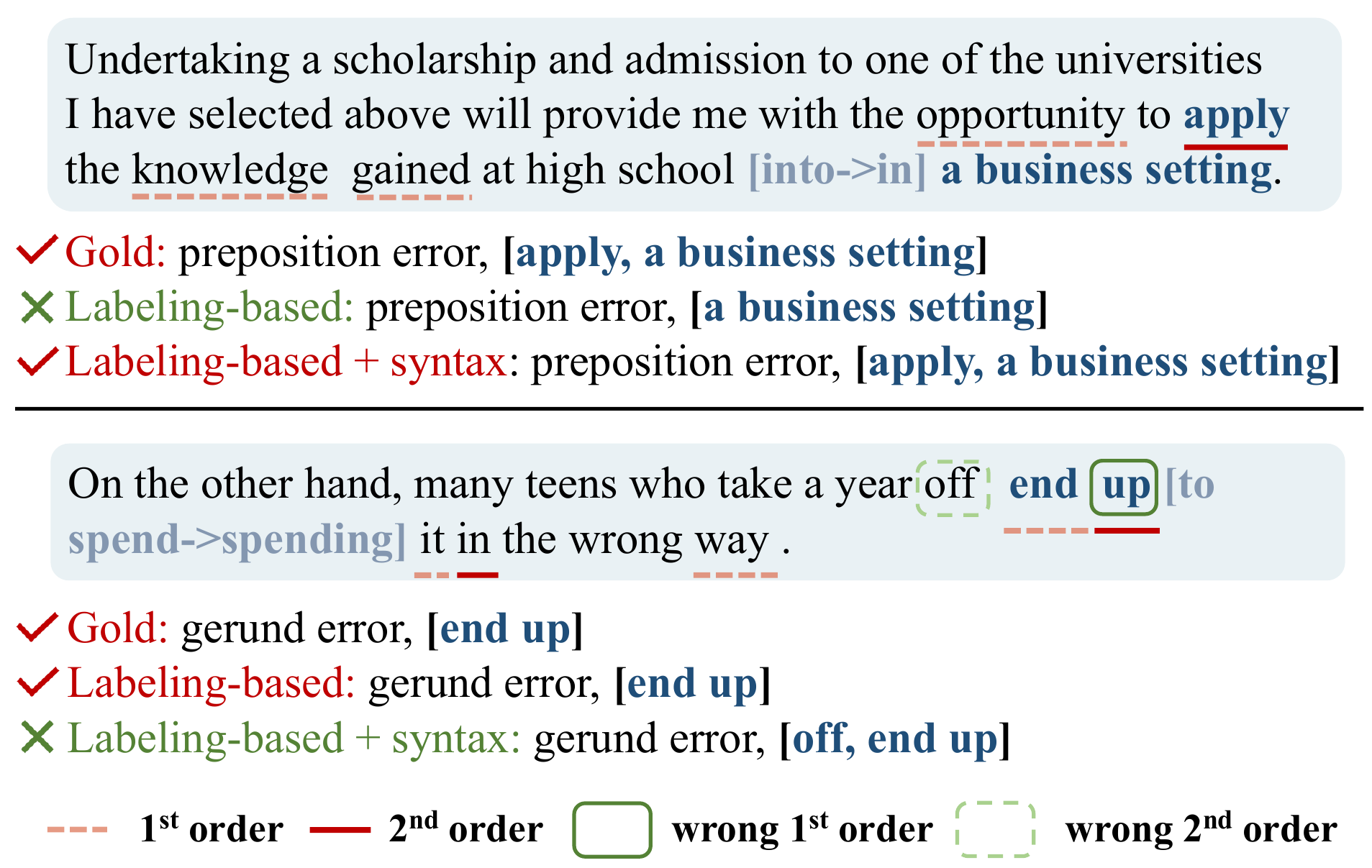}
\caption{Case study. The first case shows the identification problem for distant evidence words. The second case shows the error caused by wrong dependency parsing results.}
\label{fig:case_study}
\end{figure}

We identify two phenomena from our syntactic and non-syntactic models based on labeling models:
\paragraph{Distant Words Identification.} 
The non-syntactic model makes errors because it does not incorporate explicit syntactic modeling, particularly in long and complex sentences where it is difficult to identify distant evidence words.
As shown in the first case of Figure~\ref{fig:case_study}, the non-syntactic model fails to consider evidence words, such as ``apply'', that is located far away from the correction. However, the syntactic-based model is able to identify the evidence word ``\textit{apply}''.
\paragraph{Dependency Parsing Errors.} 
Some evidence word identification errors are from the misleading parsing results in the long sentence \cite{ma2018stack}. As shown in the second case of Figure \ref{fig:case_study}, the model with syntactic knowledge is actually using an inaccurate parse tree in the green box from the off-the-shelf parser, which results in identifying redundant word ``\textit{off}''.


\section{Conclusion}
We introduce EXPECT, an explainable dataset for grammatical error correction, which contains 21,017 instances with evidence words and error categorization annotation. We implement several models and perform a detailed analysis to understand the dataset better. Experiments show that injecting syntactic knowledge can help models to boost their performance. Human evaluation verifies the explanations provided by the proposed explainable GEC systems are effective in helping second language learners understand the corrections. We hope that EXPECT facilitates future research on building explainable GEC systems.

\section*{Limitations}
The limitations of our work can be viewed from two perspectives. Firstly, we have not thoroughly investigated seq2seq architectures for explainable GEC. Secondly, the current input of the explainable system is the gold correction during training, whereas, in practical applications, the input would be the output of a GEC system. We have not yet explored methods to bridge this gap.

\section*{Ethics Consideration}
We annotate the proposed dataset based on W\&I+LOCNESS, without copyright constraints for academic use.
For human annotation (Section~\ref{sec:human_annotatation} and Section~\ref{sec:human_eval}), we recruit our annotators from the linguistics departments of local universities through public advertisement with a specified pay rate.
All of our annotators are senior undergraduate students or graduate students in linguistic majors who took this annotation as a part-time job. We pay them 60 CNY an hour. The local minimum salary in 2022 is 25.3 CNY per hour for part-time jobs. The annotation does not involve any personally sensitive information. The annotated is required to label factual information (i.e., evidence words inside the sentence.).

\bibliography{acl2023}
\bibliographystyle{acl_natbib}

\clearpage
\appendix

\section{Appendix}
\label{sec:appendix}
\subsection{Grammatical Error Categories}
\label{sec:appendixa}
The definition of each grammatical error category in EXPECT is shown as follows: 
\begin{itemize}
\item Infinitives: including errors like missing \textit{to} before a certain verbs for to-infinitives, or unnecessary \textit{to} after modal verbs for zero-infinitives.
\item Gerund: misuse of the verb form that should act as a noun in a sentence.
\item Participles: confuse with ordinary verbs like present simple, past simple or present continuous and other participles-related situations. 
\item Subject-verb agreement(SVA): the verb didn't agree with the number of the subject.
\item Auxiliary verb: misuse of main auxiliary verbs like \textit{do}, \textit{have} or model auxiliary verbs like \textit{could}, \textit{may}, \textit{should}, etc. 
\item Verb tense: incongruities in verb tenses, such as erroneous tense shift in a compound sentence, etc.
\item Pronoun-antecedent agreement(PAA): pronouns didn't agree in number, person, and gender with their antecedents.
\item Possessive: misuse of possessive adjectives and possessive nouns.
\item Collocation: atypical word combinations that are grammatically acceptable but not common. 
\item Preposition: misuse of prepositional words. 
\item POS confusion: confusions in part of speech like noun/adjective confusion(e.g. difficulty, difficult), adjective/adverb confusion(e.g. ready, readily), etc.
\item Article: wrong use of article. 
\item Number: confusion in singular or plural form of nouns.
\item Transition: extra preposition after transitive verbs and missing proposition after intransitive verbs.
\end{itemize}

\subsection{Implementation Details}
We employ pre-trained \texttt{BERT-large-cased} in HuggingFace's Transformer  Library \cite{wolf2020transformers} as our encoder, which consists of 24 Transformer layers and 16 attention heads with 1024 hidden dimensions. We set the dimension of the correction embeddings and syntactic embeddings as 1024, which is the same as that in BERT. We set the learning rate to 1e-5 and batch size to 32 for non-interactive matrix models, and 5e-5 and 16 for interactive matrix models. 



\end{document}